\documentclass[accepted, specialissue]{melba}

\usepackage{amsmath,amsfonts,graphicx,url}
\usepackage[nolist]{acronym}
\usepackage[table]{xcolor}
\usepackage{colortbl}
\usepackage{graphicx}
\usepackage{booktabs}
\usepackage{hyphenat}
\usepackage{adjustbox}
\usepackage{rotating}
\usepackage{caption}
\usepackage{array}
\usepackage{siunitx}

\melbaid{2026:006}  
\doi{10.59275/j.melba.2026-6c1g}
\melbaauthors{Banerjee et al.}  
\email{sweta.banerjee@hs-flensburg.de}
\volume{2026}
\firstpageno{115}  
\melbayear{2026}  
\datesubmitted{2025-07-15}  
\datepublished{2026-03-12}  

\melbaspecialissue{MELBA–BVM 2025 Special Issue}
\melbaspecialissueeditors{Andreas Maier, Thomas Deserno, Heinz Handels, Klaus Maier-Hein, Christoph Palm, Thomas Tolxdorff, Katharina Breininger}

\ShortHeadings{Benchmarking Atypical Mitosis Classification}{Banerjee et al.}

\title{Benchmarking Deep Learning and Vision Foundation Models for Atypical vs. Normal Mitosis Classification with Cross-Dataset Evaluation}

\author{
    \firstname Sweta \surname Banerjee\aff{1}\orcid{0000-0001-5029-5378},
    \firstname Viktoria \surname Weiss\aff{2}\orcid{0009-0002-0062-844X},
    \firstname Taryn A. \surname Donovan\aff{3}\orcid{0000-0001-5740-9550},
    \firstname Rutger H.J. \surname Fick\aff{4},
    \firstname Thomas \surname Conrad\aff{5},
    \firstname Jonas \surname Ammeling\aff{6}\orcid{0000-0002-0335-1194},
    \firstname Nils \surname Porsche\aff{1},
    \firstname Robert \surname Klopfleisch\aff{5}\orcid{0000-0002-6308-0568},
    \firstname Christopher C. \surname Kaltenecker\aff{7}\orcid{0000-0002-7795-2726},
    \firstname Katharina \surname Breininger\aff{8}\orcid{0000-0001-7600-5869},
    \firstname Marc \surname Aubreville\aff{1}\orcid{0000-0002-5294-5247},
    \firstname Christof A. \surname Bertram\aff{2}\orcid{0000-0002-2402-9997}
}
\affiliations{
    \num 1 \addr Flensburg University of Applied Sciences, Germany \\
    \num 2 \addr University of Veterinary Medicine, Vienna, Austria \\
    \num 3 \addr The Schwarzman Animal Medical Center, New York, USA \\
    \num 4 \addr Diffusely, Paris, France \\
    \num 5 \addr Freie Universit\"{a}t Berlin, Berlin,  Germany \\
    \num 6 \addr Technische Hochschule Ingolstadt, Ingolstadt, Germany \\
    \num 7 \addr Medical University of Vienna, Austria \\
    \num 8 \addr Julius-Maximilians-Universit\"{a}t W\"{u}rzburg, W\"{u}rzburg, Germany
}

\abstract{
	Atypical mitosis marks a deviation in the cell division process that has been shown be an independent prognostic marker for tumor malignancy. However, atypical mitosis classification remains challenging due to low prevalence, at times subtle morphological differences from normal mitotic figures, low inter-rater agreement among pathologists, and class imbalance in datasets. Building on the Atypical Mitosis dataset for Breast Cancer (AMi-Br), this study presents a comprehensive benchmark comparing deep learning approaches for automated atypical mitotic figure (AMF) classification, including end-to-end fine-tuned deep learning models, foundation models with linear probing, and foundation models fine-tuned with low-rank adaptation (LoRA). For rigorous evaluation, we further introduce two new held-out AMF datasets - AtNorM-Br, a dataset of mitotic figures from the TCGA breast cancer cohort, and AtNorM-MD, a multi-domain dataset of mitotic figures from a subset of the MIDOG++ training set. We found average balanced accuracy values of up to 0.8135, 0.7788, and 0.7723 on the in-domain AMi-Br and the out-of-domain AtNorm-Br and AtNorM-MD datasets, respectively. Our work shows that atypical mitotic figure classification, while being a challenging problem, can be effectively addressed through the use of recent advances in transfer learning and model fine-tuning techniques. We make all code and data used in this paper available in this github repository: \url{https://github.com/DeepMicroscopy/AMi-Br_Benchmark}.}


\begin{document}

\twocolumn[\maketitle]

\section{Introduction}

\begin{figure*}[t]
    \centering
    \includegraphics[width=\linewidth]{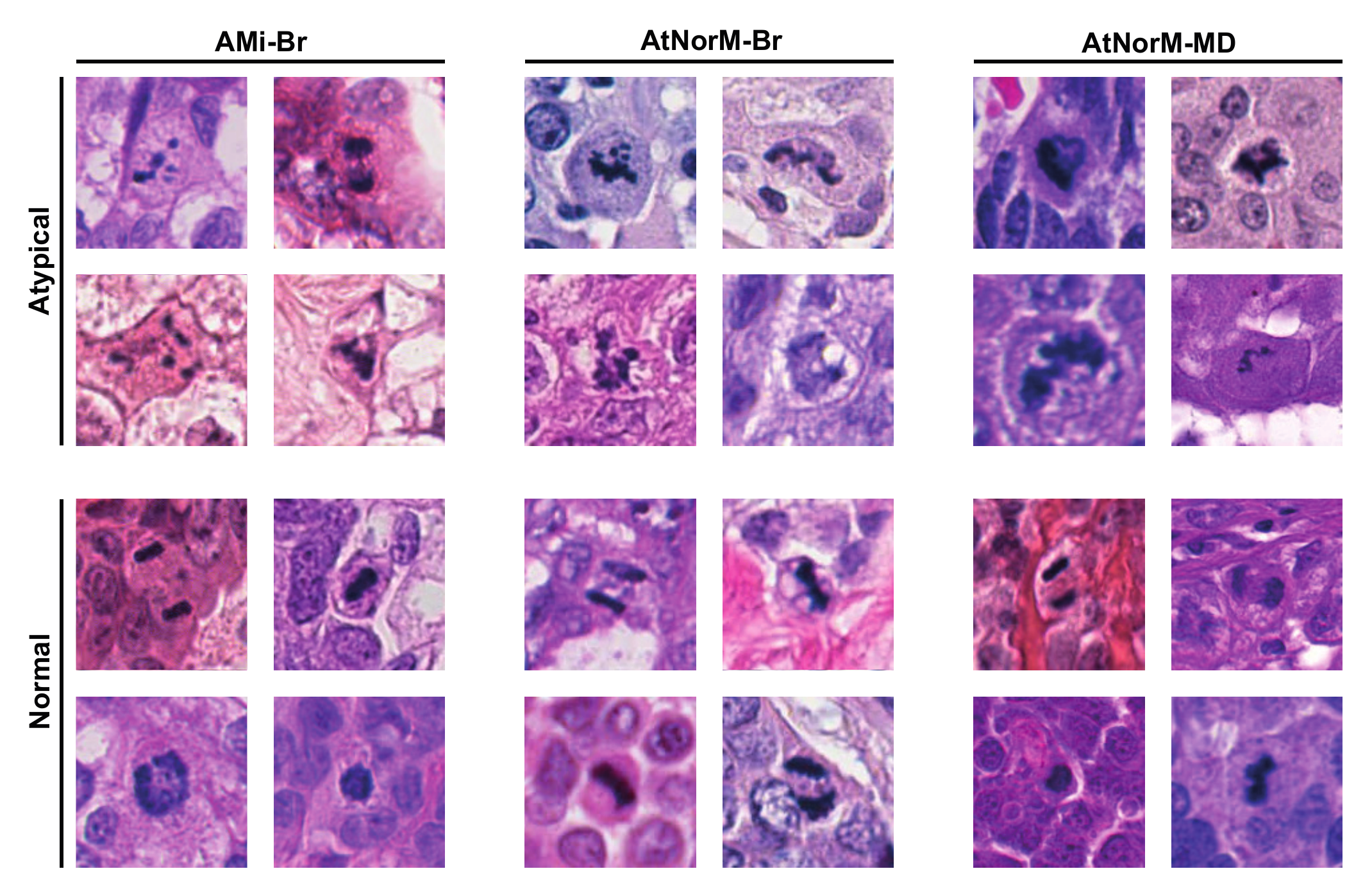}
    \caption{Image samples showing atypical and normal mitotic figures from all three datasets included in this study}
    \label{fig:mitosis-image-samples}
\end{figure*}

Mitosis is the process where a cell replicates its genetic material (DNA) and then divides into two identical daughter cells. This process enables the proliferation of cells, supporting physiological tissue growth and facilitating the replacement of old, damaged, or worn-out cells. The dividing cell may be observed microscopically as a \ac{MF} with specific morphologies resembling the highly regulated cell division phases. However, cell proliferation does not occur only under physiological conditions; it is also a hallmark of cancer, where cell division is dysregulated and increased. Subsequently, quantifying cell proliferation is essential for estimating the aggressiveness of many human and animal cancers, including human breast cancer~\citep{fitzgibbons2023protocol,kiupel2011proposal,Louis2016}. The routine method in the assessment of the growth rate is the mitotic count, which reflects the number of \acp{MF} within a specific tumor area, selected from a \ac{WSI}. Elevated mitotic counts are associated with an increased risk of distant metastasis and higher risk of death~\citep{medri2003prognostic}.

In tumor samples, two variants of \acp{MF} can be found: normal and atypical.
Normal \acp{MF} indicate the orderly process of cell division, designated by the prometaphase, metaphase, anaphase, and telophase, where chromosomes align and separate equally to produce two identical daughter cells~\citep{donovan2021mitotic}. An \ac{AMF}, on the other hand, is a cell undergoing an abnormal cell division process characterized by errors in chromosome segregation or other morphological irregularities, which promote cancer progression~\citep{donovan2021mitotic}. Until recently, a correlation of the count of \acp{AMF} with shorter patient survival has been demonstrated in only few studies on breast cancer~\citep{ohashi2018prognostic,lashen2022characteristics} and selected other tumor types~\citep{jin2007distinct,kalatova2015tripolar,bertram2023atypical,matsuda2016mitotic}. 
A very recent study by \cite{jahanifar2025pan} has expanded these insights and the corresponding prognostic relevance of the rate of \acp{AMF} across a wide range of tumors in a large-scale evaluation, highlighting the importance of further investigations of this prognostic parameter. 

Identification of \acp{MF} is traditionally done by pathologists, making it highly subjective and time intensive. This process is further complicated by inter-observer variability, low prevalence of \acp{MF}, and inconsistencies in slide quality~\citep{Meyer:2005cl,Meyer:2009eu,wilm2021influence}. To address these challenges, deep learning-based \ac{MF} detection has been extensively investigated~\citep{aubreville2024domain, veta2019predicting, aubreville2023mitosis}, demonstrating a promising outlook for improving both accuracy and efficiency and reducing the diagnostic workload for pathologists. Along similar lines, deep learning models offer potential for the automatic differentiation of \acp{AMF} against normal \acp{MF}.
For instance, this could enable the quantification of the ratio of \acp{AMF} across entire whole slide images~\citep{jahanifar2025pan}, which is otherwise impractical to perform manually due to time constraints and the rarity of \acp{AMF}. 
Such quantitative assessments can support more objective and reproducible evaluations of tumor aggressiveness, potentially improving treatment planning in clinical settings.

At the time of this writing, there have been very few studies investigating deep learning for \acp{AMF}~\citep{aubreville2023deep, fick2024improving, bertram2025histologic}. The automated classification of atypical vs.\ normal \acp{MF} is complicated by the substantial variation of morphologies, often with some mitotic figures displaying overlapping features of both normal and \acp{AMF} \citep{donovan2021mitotic,bertram2023atypical}, and the low frequency of \acs{AMF}, resulting in class imbalance. 
Given the prognostic relevance of differentiating normal vs.\ atypical \acp{MF}, standardized benchmarks, such as publicly available datasets, evaluation protocols, and baseline models, may serve as facilitators to push further developments in this field that tackle these challenges. 
In this context, foundation models offer a promising direction. These models, typically \ac{ViT}-based and pre-trained with self-supervised learning on millions of histopathology tiles, have shown strong performance across various medical imaging tasks, including classification~\citep{campanella2025clinical, breen2025comprehensive}. When subsequently fine-tuned on comparatively small, task-specific labelled datasets, they are particularly effective in small data regimes. These models are typically employed either via linear probing, where a lightweight classifier is trained on top of a frozen backbone, or through parameter-efficient fine-tuning techniques like \ac{LoRA} \citep{hu2022lora}. However, comprehensive studies leveraging these techniques for atypical vs.\ normal mitosis classification are still lacking.

The contribution of this work is two-fold: First, we present three datasets, designed for the automated identification of \acp{AMF} to establish suitable datasets that reflect the unique challenges of this task. Second, we provide a comprehensive benchmark comparing end-to-end fine-tuned deep learning architectures, foundation models with linear probing, and foundation models fine-tuned with \ac{LoRA} for the task of atypical versus normal mitotic figure classification. 


\section{Datasets}

We present and use three distinct datasets in this paper - AMi-Br, AtNorM-Br, and AtNorM-MD as described below. This work is an extension of a previous conference contribution by our group \citep{bertram2025histologic}, which already introduced the first of those datasets (AMi-Br) in brief. Representative image samples of atypical and normal mitotic figures from all three datasets are shown in Figure~\ref{fig:mitosis-image-samples}.

Being the largest of the three datasets used in this study, we use data from the AMi-Br dataset for training in all evaluations. This allows us to use the other two newly introduced datasets as independent held-out sets. It has been shown that data distribution shifts (domain shifts) impede the performance of deep learning models considerably \citep{stacke2020measuring,aubreville2021quantifying}. With these datasets, we expect varying degrees of domain shifts, and which allows us to investigate generalization performance. 
Only a part of the AMi-Br dataset is used for training the models for binary classification of atypical vs.\ normal mitoses, and the rest is used as another held-out test dataset. 
The dataset statistics, including total number of atypical and normal \acp{MF} in each, annotation type, source dataset(s), species involved etc.\ have been summarized in Table~\ref{tab:dataset-stats}.

\begin{table}[t]
    \centering
    \caption{Dataset statistics including sample counts, class balance, annotation details, and domain coverage.}
    \label{tab:dataset-stats}
    \resizebox{\columnwidth}{!}{%
    \begin{tabular}{lccc}
        \toprule
        \textbf{Property} & \textbf{AMi-Br} & \textbf{AtNorM-Br} & \textbf{AtNorM-MD} \\
        \midrule
        Total \acp{MF}        & 3,720 & 746  & 2,107 \\
        Atypical              & 832   & 128  & 219   \\
        Normal                & 2,888 & 618  & 1,888 \\
        \ac{AMF} Rate (\%)    & 22.4\% & 17.2\% & 10.4\% \\
        Annotation Type       & 3-expert vote & Single expert & 5-expert vote \\
        Expert Agreement      & 78.2\% & --   & 69.6\% \\
        Source Dataset(s)     & TUPAC16, MIDOG21 & TCGA (BRCA) & MIDOG++ \\
        Species               & Human & Human & Human + Canine \\
        \bottomrule
    \end{tabular}%
    }
\end{table}

\subsection{AMi-Br}
The first dataset, AMi-Br~\citep{bertram2025histologic} consists of a total of 3,720 \acp{MF} from human breast cancer, of which 1,999 \acp{MF} are from the TUPAC16~\citep{veta2019predicting} alternative label set \citep{bertram2020pathologist} and 1,721 \acp{MF} are from the MIDOG21 training dataset~\citep{aubreville2023mitosis}. The \acp{MF} were graded into one of two classes --- atypical or normal --- by three expert pathologists, two of whom were board-certified. According to the result of the majority vote between the three experts, 832 \acp{MF} were found to be atypical, while the remaining 2,888 were classified as normal \acp{MF}. We found total agreement per object by all three experts in 2908 (78.2\%) of the cases. 

\subsection{AtNorM-Br}
The third dataset, \ac{AtNorM}-Br, also made available publicly within the scope of this work, contains 746 \ac{MF} instances from 179 patients from the breast cancer (BRCA) cohort of the \ac{TCGA} \citep{lingle2016cancer}. \ac{TCGA} contains images from various sources and with partially mixed quality, stemming from differences in staining protocols, scanning equipment, and tissue preparation across institutions, and is thus also a valuable asset for the assessment of generalization. This dataset was annotated by a single expert with high experience in \ac{AMF} classification, according to which 128 \acp{MF} were found to be atypical and the remaining 618 were classified as normal, yielding an \ac{AMF} rate of 17.16\%.

\subsection{AtNorM-MD}
The second dataset, \ac{AtNorM}-MultiDomain (MD) extends beyond human breast cancer and covers six domains, spanning both human and canine tumors. These include canine lung cancer, canine lymphoma, canine cutaneous mast cell tumor, human neuroendocrine tumor, canine soft tissue sarcoma, and human melanoma. It was sampled randomly from all but the human breast cancer domains of the publicly available MIDOG++ dataset~\citep{aubreville2023comprehensive}. It comprises 2,107 \acp{MF} from 70 patients, with 219 (10.4\%) being atypical. Labeling was performed as a majority vote of five pathology experts, three of whom were board-certified. We found total agreement per object by all five experts in 1,466 (69.6\%) of the cases.

\begin{table}[t]
    \centering
    \scriptsize
    \caption{Overview of foundation models used in this paper}
    \label{tab:fm}
    \begin{adjustbox}{max width=\linewidth}
    \begin{tabular}{l c c c c} 
        \toprule
        \textbf{Model} & \multicolumn{2}{c}{\textbf{Training Dataset Size}} & \textbf{Model Type} & \textbf{Size (Params)} \\
        \cmidrule(lr){2-3}
        \textbf{} & \textbf{WSIs} & \textbf{Tiles} & \textbf{} & \textbf{} \\
        \midrule
        UNI & 100K & 100M & ViT-L/16 & 307M \\
        UNI2-h & 350K & 200M & ViT-H/14 & 681M \\
        Virchow & 1.5M & 2B & ViT-H/14 & 632M \\
        Virchow2 & 3.4M & 1.7B & ViT-H/14 & 632M \\
        Prov-Gigapath & 170K & 1.3B & ViT-g/16 & 1.1B \\
        H-Optimus-0 & 500K & - & ViT-g/14 & 1.1B \\
        H-Optimus-1 & 1M & 2B & ViT-g/14 & 1.13B \\
        \bottomrule
    \end{tabular}
    \end{adjustbox}
\end{table}

\section{Methods}
\subsection{End-to-end fine-tuned Baseline Deep Learning Models}
To establish robust baselines for atypical vs.\ normal mitoses classification, we evaluate three widely-used deep learning architectures -- EfficientNetV2~\citep{tan2021efficientnetv2}, \ac{ViT}~\citep{dosovitskiy2020image}, and Swin Transformer~\citep{liu2021swin}. All of these models have demonstrated strong performance across various medical imaging datasets and tasks. EfficientNetV2 is a well-established \ac{CNN} known for its strong performance and efficiency in various classification tasks. It serves as a representative of traditional \ac{CNN}-based approaches. In contrast, the \ac{ViT} represents a major shift from convolutional architectures to transformer-based models. A \ac{ViT} processes images as sequences of patches and uses self-attention mechanisms to model global relationships across the input. This global receptive field allows the \ac{ViT} to capture long-range dependencies, which can be particularly useful for complex histopathological patterns. The Swin Transformer further advances the transformer-based approach by introducing a hierarchical architecture with shifted windows. This architecture enables the model to compute attention locally while gradually building up to global representations, providing a balance between computational efficiency and the ability to capture both local details and global structure. Swin Transformers have been particularly effective in dense prediction tasks and high-resolution image analysis, making it well-suited for mitotic figure classification \citep{liu2021swin}. The selected model architectures were chosen to be of similar size in parameters. Each model was fine-tuned in an end-to-end manner and evaluated on three held-out test datasets: a random, patient-stratified split of the AMi-Br dataset, accounting for $\approx$ 22 \% of the samples in the dataset (samples not used in training or validation), and the entire AtNorM-MD and AtNorM-Br datasets. 

\subsection{Foundation Models}

Foundation models are models that are trained on large and diverse corpora of data, often using unsupervised techniques like self supervision. The goal of foundation models is to train feature extractors that generalize well and are easily adaptable to down-stream tasks. In the field of computational pathology, we have seen a very strong incline towards publicly available foundation models, calling for an investigation on the task of \ac{AMF} classification.
We compare eight state-of-the-art models -- UNI~\citep{chen2024towards}, UNI2-h~\citep{chen2024towards, mahmoodlab_uni2h_2025}, Virchow~\citep{vorontsov2024foundation}, Virchow2~\citep{zimmermann2024virchow2}, Prov-Gigapath~\citep{xu2024whole}, H-Optimus-0~\citep{hoptimus0}, H-Optimus-1~\citep{hoptimus1}, and H0-mini~\citep{filiot2025distillingfoundationmodelsrobust}. An overview of selected details for each model such as training dataset size, size in parameters and model type for training is provided in Table~\ref{tab:fm}. All selected foundation models are \ac{ViT}-based and have been pre-trained using the DINOv2 algorithm \citep{oquab2023dinov2} on datasets comprising of hundreds of thousands to multiple millions of \acp{WSI}.

\subsubsection{Linear Probing of Foundation Models}
We use a linear probing strategy, where the feature extractor (i.e., the foundation model) is used to extract the embeddings. After that, the foundation model is kept frozen, and a linear classifier is trained on top of the extracted features. This process helps us to assess the ability of the representations of the foundation models to classify atypical vs.\ normal \acp{MF}.

\subsubsection{LoRA Fine-Tuning}
Parameter-Efficient Fine-Tuning (PEFT) methods~\citep{houlsby2019parameter} are model adaptation strategies which focus on fine-tuning models with minimal changes to their parameters.
Traditional fine-tuning updates all model parameters, which becomes computationally expensive for large models. 
\ac{LoRA}~\citep{hu2022lora}, a PEFT method, addresses this by freezing the original weights and introducing small, trainable low-rank matrices into specific layers, typically within the query, key, and value weight matrices of self-attention blocks. Only these matrices are updated during training, greatly reducing memory and compute requirements while retaining performance comparable to full fine-tuning. We apply \ac{LoRA}-based finetuning to the same set of models previously evaluated with linear probing -- UNI, UNI2-h, Virchow, Virchow2, Prov-Gigapath, H-Optimus-0 and H-Optimus-1, for a direct comparison.

\begin{figure*}[t]
    \centering
    \includegraphics[width=\linewidth]{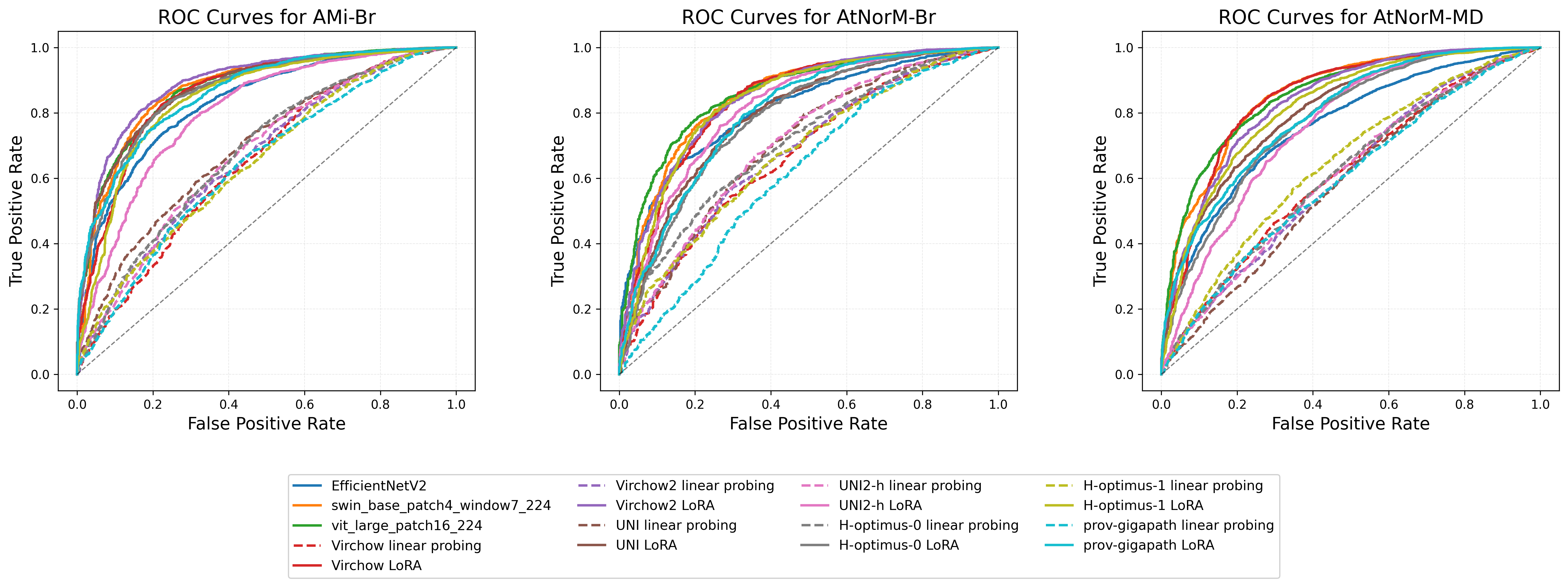}
    \caption{\Acf{ROC} curves for baseline, foundation models linear probing and LoRA finetuning of foundation models across three datasets.}
    \label{fig:myfigure}
\end{figure*}

\section{Experimental Setup}

We set up detailed experiments for the above cases. All training was performed exclusively on the AMi-Br dataset. The image preprocessing for the binary classification experiments included resizing all patches, originally of the size $128 \times 128$, to $224 \times 224$ pixels. Training augmentations consisted of random horizontal flip, rotation, color jitter, and random resized crop, followed by normalization. For models initialized from ImageNet-pretrained weights and fine-tuned end-to-end, we used the standard ImageNet mean and standard deviation for normalization. For the computational pathology foundation models used in the linear probing and \ac{LoRA} experiments, we instead followed the pre-processing recommended for each model, i.e., normalizing with the model-specific mean and standard deviation used during their pre-training, rather than assuming ImageNet statistics. Validation images were only resized and normalized. The best checkpoint for each fold was selected based on validation balanced accuracy. The balanced accuracy for binary classification is the average of sensitivity and specificity and can be mathematically stated as follows:
\[
\text{Balanced Accuracy} = \frac{\text{Sensitivity} + \text{Specificity}}{2}
\]The balanced accuracy used as the primary metric because of the class imbalance between the atypical and normal \acp{MF} in the dataset, with a fixed decision threshold of 0.5 on the predicted probabilities. All experiments were performed using 5-fold cross validation, where stratification of the data was based on the slide (patient) to ensure that all patches from a given slide appeared in only one split (train or validation), thus preventing data leakage. To address class imbalance during training, we used a weighted random sampler that assigns higher sampling weights to under-represented classes and lower weights to frequent classes, with each sample’s weight set inversely to its class frequency. We used weighted random sampling because it was proven to give the best results in previous work \citep{bertram2025histologic}.
 All models were trained using the Adam optimizer with L2 regularization (learning rate: $5 \times 10^{-5}$, weight decay: $1 \times 10^{-5}$) and cross-entropy loss. We applied early stopping with a patience of 15 epochs based on validation balanced accuracy, and used learning rate scheduling on plateau (factor: 0.5, patience: 3, minimum LR: $1 \times 10^{-7}$). Each model was trained for up to 100 epochs with a batch size of 8. The rest of the case-specific details are described below in the subsections.

\subsection{End-to-end fine-tuned Baseline Deep Learning Models}

We initialized all three models (i.e., EfficientNetV2, \ac{ViT} and Swin Transformer) that were fine-tuned in an end-to-end fashion, essentially doing a full finetuning of all the layers with ImageNet-pretrained weights, and the original classification head was replaced with a custom binary classifier consisting of a single linear layer. Specifically, we used the \texttt{vit\_large\_patch16\_224} model from the \texttt{timm} library \citep{rw2019timm}, a ViT with 224×224 input resolution, 16×16 patch size, and 86.6 million parameters. The Swin Transformer variant used was \texttt{swin\_base\_patch4\_window7\_224} which features a 4×4 patch size and 7×7 shifted attention windows, comprising 87.8 million parameters. For EfficientNetV2, we employed \texttt{efficientnetv2\_m} from \texttt{timm}, which has approximately 54.1 million parameters. This protocol was applied identically to all three models to enable fair comparison of their performance on the atypical mitosis classification task.

\subsection{Linear Probing with Foundation Models}

We conducted comprehensive evaluation of seven foundation models using a standardized linear probing method for atypical versus normal mitosis classification. For each model, we extracted high-dimensional feature embeddings from histopathology images and trained a single linear classification layer while keeping the pre-trained feature extractor frozen. This method was consistently applied across all six foundation models to ensure fair comparison of their feature representation capabilities on the atypical classification task.

\subsection{LoRA Fine-tuning of Foundation Models}

For the \ac{LoRA}-based fine-tuning of the foundation models, each model was initialized with publicly available pre-trained weights and adapted using LoRA with rank 8, scaling factor 16, and dropout rate 0.3, applied to both the transformer attention layers (i.e., query, key, value, and output projections) and the MLP components (i.e., feedforward layers fc1 and fc2) of the transformer blocks. The classification head was re-initialized and trained jointly with \ac{LoRA} modules, while the rest of the backbone remained frozen. This standardized protocol ensured fair and rigorous evaluation of LoRA-based adaptation across diverse foundation models on the AMi-Br benchmark.

\begin{table*}[t]
    \centering
    \scriptsize
        \caption{Classification performance of the evaluated models on the AMi-Br, AtNorM-Br, and AtNorM-MD test sets. For each dataset, balanced accuracy and AUROC are reported as mean $\pm$ standard deviation across cross-validation folds. All models were trained exclusively on the AMi-Br dataset, while evaluation was performed on the respective test sets. Within each column, bold values denote the best-performing model for that dataset and  in the respective group of models.}
    \label{tab:experiments-3}
    \begin{adjustbox}{max width=\linewidth}
    \begin{tabular}{l*{3}{cc}}
        \toprule
        \textbf{Model} & 
        \multicolumn{2}{c}{\textbf{AMi-Br}} & 
        \multicolumn{2}{c}{\textbf{AtNorM-Br}} & 
        \multicolumn{2}{c}{\textbf{AtNorM-MD}} \\
        \cmidrule(lr){2-3} \cmidrule(lr){4-5} \cmidrule(lr){6-7}
        & \textbf{Balanced Acc.} & \textbf{AUROC} 
        & \textbf{Balanced Acc.} & \textbf{AUROC} 
        & \textbf{Balanced Acc.} & \textbf{AUROC} \\
        \midrule
        EfficientNetV2 (Baseline) & $0.7474 \pm 0.0167$ & $0.8315 \pm 0.0168$ & $0.7283 \pm 0.0168$ & $0.8110 \pm 0.0084$ & $0.6975 \pm 0.0277$ & $0.7604 \pm 0.0217$ \\
        Vision Transformer (Baseline) & $0.7899 \pm 0.0268$ & $0.8872 \pm 0.0107$ & $\textbf{0.7788} \pm 0.0229$ & $\textbf{0.8710} \pm 0.0075$ & $0.7494 \pm 0.0342$ & $0.8720 \pm 0.0117$ \\
        Swin Transformer (Baseline) & $\textbf{0.8052} \pm 0.0124$ & $\textbf{0.9029} \pm 0.0112$ & $0.7752 \pm 0.0134$ & $0.8683 \pm 0.0113$ & $\textbf{0.7723} \pm 0.0142$ & $\textbf{0.8806} \pm 0.0087$ \\
        \midrule
        UNI & $\textbf{0.6339} \pm 0.0189$ & $\textbf{0.6949} \pm 0.0275$ & $0.6406 \pm 0.0339$ & $0.7028 \pm 0.0310$ & $0.5510 \pm 0.0123$ & $0.5892 \pm 0.0118$ \\
        UNI2-h & $0.6220 \pm 0.0095$ & $0.6765 \pm 0.0157$ & $\textbf{0.6482} \pm 0.0214$ & $0.7059 \pm 0.0230$ & $0.5773 \pm 0.0212$ & $0.6119 \pm 0.0357$ \\
        Virchow & $0.6029 \pm 0.0132$ & $0.6520 \pm 0.0168$ & $0.5969 \pm 0.0155$ & $0.6733 \pm 0.0266$ & $0.5623 \pm 0.0235$ & $0.6184 \pm 0.0566$ \\
        Virchow2 & $0.6089 \pm 0.0192$ & $0.6637 \pm 0.0214$ & $0.6236 \pm 0.0296$ & $0.6795 \pm 0.0344$ & $0.5694 \pm 0.0173$ & $0.6046 \pm 0.0324$ \\
        Prov-Gigapath & $0.6097 \pm 0.0177$ & $0.6435 \pm 0.0188$ & $0.5893 \pm 0.0074$ & $0.6187 \pm 0.0193$ & $0.5558 \pm 0.0125$ & $0.6039 \pm 0.0244$ \\
        H-Optimus-0 & $0.6319 \pm 0.0153$ & $0.6801 \pm 0.0181$ & $0.6369 \pm 0.0147$ & $\textbf{0.7090} \pm 0.0163$ & $0.5821 \pm 0.0195$ & $0.6142 \pm 0.0207$ \\ 
        H-Optimus-1 & $0.5943 \pm 0.0132$ & $0.6518 \pm 0.0152$ & $0.6271 \pm 0.0168$ & $0.6777 \pm 0.0275$ & $\textbf{0.5966} \pm 0.0317$ & $\textbf{0.6462} \pm 0.0461$ \\ 
        \midrule
        UNI (LoRA) & $0.7952 \pm 0.0092$ & $0.8839 \pm 0.0059$ & $0.7183 \pm 0.0226$ & $0.7979 \pm 0.0142$ & $0.7069 \pm 0.0278$ & $0.8222 \pm 0.0224$ \\
        UNI2-h (LoRA) & $0.7138 \pm 0.0121$ & $0.8153 \pm 0.0211$ & $0.7301 \pm 0.0152$ & $0.8228 \pm 0.0143$ & $0.6914 \pm 0.0321$ & $0.7616 \pm 0.0415$ \\
        Virchow (LoRA) & $0.7878 \pm 0.0250$ & $0.8891 \pm 0.0150$ & $\textbf{0.7696} \pm 0.0198$ & $0.8540 \pm 0.0200$ & $\textbf{0.7705} \pm 0.0287$ & $\textbf{0.8641} \pm 0.0247$ \\
        Virchow2 (LoRA) & $\mathbf{0.8135} \pm 0.0145$ & $\textbf{0.9026} \pm 0.0051$ & $0.7632 \pm 0.0190$ & $\textbf{0.8579} \pm 0.0117$ & $0.7424 \pm 0.0305$ & $0.8503 \pm 0.0171$ \\
        Prov-Gigapath (LoRA) & $0.7602 \pm 0.0113$ & $0.8682 \pm 0.0122$ & $0.7263 \pm 0.0296$ & $0.8077 \pm 0.0184$ & $0.7007 \pm 0.0228$ & $0.8073 \pm 0.0259$ \\
        H-Optimus-0 (LoRA) & $0.7846 \pm 0.0169$ & $0.8721 \pm 0.0148$ & $0.7044 \pm 0.0383$ & $0.7921 \pm 0.0271$ & $0.6549 \pm 0.0281$ & $0.7879 \pm 0.0163$ \\
        H-Optimus-1 (LoRA) & $0.7762 \pm 0.0110$ & $0.8611 \pm 0.0083$ & $0.7658 \pm 0.0298$ & $0.8524 \pm 0.0100$ & $0.7396 \pm 0.0202$ & $0.8389 \pm 0.0104$ \\
        \bottomrule
    \end{tabular}
    \end{adjustbox}
\end{table*}

\begin{table}[t]
\centering
\caption{\textit{p}-values for group comparisons of balanced accuracies within each dataset.}
\label{tab:stat_tests_pvalues}
\setlength{\tabcolsep}{3.5pt}
\renewcommand{\arraystretch}{1.15}

\resizebox{\linewidth}{!}{%
\begin{tabular}{l S[table-format=1.2e-2] S[table-format=1.2e-2] S[table-format=1.3]}
\toprule
Dataset &
{\shortstack{Linear Probing vs LoRA\\(Wilcoxon signed-rank)}} &
{\shortstack{Baseline vs Linear Probing\\(Mann--Whitney U)}} &
{\shortstack{Baseline vs LoRA\\(Mann--Whitney U)}} \\
\midrule
AMi-Br    & 5.82e-11 & 2.91e-08 & 0.73 \\
AtNorM-Br & 1.16e-10 & 2.91e-08 & 0.06 \\
AtNorM-MD & 5.82e-11 & 2.91e-08 & 0.09 \\
\bottomrule
\end{tabular}%
}

\vspace{2pt}
\end{table}

\section{Results}

We now present the comparative performance of all models across the AMi-Br, AtNorM-Br, and AtNorM-MD datasets using balanced accuracy and AUROC as primary metrics. The full quantitative results are summarized in Table~\ref{tab:experiments-3}, where we report the mean and standard deviation on the hold-out datasets across five train-validation folds of the training part of AMi-Br. To complement the tabulated results, Figure~\ref{fig:myfigure} illustrates the corresponding \ac{ROC} curves for each dataset, comparing baseline models, foundation models with linear probing, and those fine-tuned using \ac{LoRA}. 

On the AMi-Br dataset, the best-performing model in terms of balanced accuracy was Virchow2 with \ac{LoRA} finetuning, achieving an average balanced accuracy of 0.8135 (sensitivity = 0.7511, specificity = 0.8760 of the atypical class). While the highest AUROC of 0.9029 was obtained by the Swin Transformer model, Virchow2 closely followed with an AUROC of 0.9026, making it the most balanced performer overall. Other LoRA-fine-tuned models such as UNI (0.7952 mean balanced accuracy), Virchow (0.7878 mean balanced accuracy) and H-Optimus-0 (0.7846 mean balanced accuracy) also demonstrated strong performance, significantly improving over their respective linear probing, as demonstrated by the pairwise Wilcoxon signed-rank tests for each test dataset (see Tab. \ref{tab:stat_tests_pvalues}). Among the end-to-end fine-tuned baseline models, the Swin Transformer achieved the highest balanced accuracy (0.8052) and AUROC (0.9029), outperforming both EfficientNetV2 and \ac{ViT}. Foundation models with just linear probing lagged behind, with balanced accuracies ranging between 0.60-0.65.

On the AtNorM-Br dataset, the end-to-end fine-tuned \ac{ViT} was the best performing model overall with a mean balanced accuracy of 0.7788 (sensitivity = 0.7109, specificity = 0.8466 of the atypical class) and AUROC 0.8710, while the Swin Transformer followed closely with a mean balanced accuracy of 0.7752 and an AUROC of 0.8683. Among the \ac{LoRA}-fine-tuned foundation models, Virchow performed the best in terms of a mean balanced accuracy of 0.7696, followed by H-Optimus-1 (0.7658 mean balanced accuracy) and Virchow2 (0.7632 mean balanced accuracy) Other \ac{LoRA}-fine-tuned models such as UNI2-h (0.7301 balanced accuracy), Prov-Gigapath (0.7263 balanced accuracy), UNI (0.7183 balanced accuracy) and H-Optimus-0 (0.7044 balanced accuracy) also showed consistent improvements over their linear probing variants. Foundation models with linear probing performed poorly, with balanced accuracies ranging between 0.59 and 0.65.

The AtNorM-MD dataset, which represents a distinct distribution shift, saw a general drop in performance across models. The end-to-end fine-tuned Swin Transformer performed the best, with a mean balanced accuracy of 0.7723 (sensitivity = 0.6548, specificity = 0.8897 of the atypical class) and an AUROC of 0.8806. Virchow pretrained with \ac{LoRA} followed closely, with a balanced accuracy of 0.7705. Other \ac{LoRA}-fine-tuned models also performed well, including Virchow2 (0.7424 balanced accuracy), H-Optimus-1 (0.7396 balanced accuracy), UNI (0.7069 balanced accuracy), Prov-Gigapath (0.7007 balanced accuracy), UNI2-h (0.6914 balanced accuracy) and H-Optimus-0 (0.6549 balanced accuracy), all of which substantially outperformed their linear probing counterparts. Among the end-to-end fine-tuned baselines, \ac{ViT} followed closely after Swin Transformer, achieving 0.7494 balanced accuracy and 0.8720 AUROC. EfficientNetV2 achieved the least balanced accuracy of 0.6975 among the baselines. As with the other datasets, the foundation models without \ac{LoRA} adaptation performed less robustly, with balanced accuracies ranging between 0.55 and 0.60. 


\section{Discussion}

Our evaluation confirmed that all foundation models exhibited some degree of generalization to the \ac{AMF} identification task, but the extent of their effectiveness varied considerably. The application of \ac{LoRA} fine-tuning revealed particularly striking performance differences across models. Also, it is worthwhile to note that across nearly all datasets and performance metrics (balanced accuracy and AUROC), there consistently exists at least one end-to-end fine-tuned baseline model that outperforms both linear probing and LoRA-adapted foundation models.

We attribute this to not sufficiently discriminatory features extracted by the foundation models, of which many were not even trained on high magnification ($40\times$) patches \citep{zimmermann2024virchow2}, leading to a considerable data distribution shift. Thus, the stronger adaptation of the model (by using rank-reduced adaptation through LoRA) provides considerable benefit for improving the extracted features, while using the full capacity of the model (full fine-tuning) yields even better results.

Most importantly, the primary scope of our study was not to identify a single best-performing model for the respective datasets for the task of \ac{AMF} classification, but rather to compare different learning strategies for \ac{AMF} classification - full fine-tuning of ImageNet-pretrained models, \ac{LoRA}-based adaptation of foundation models, and linear probing of foundation models. To investigate this research question in a rigorous way, we performed statistical tests across all datasets to assess whether the performance differences between these learning strategies are statistically significant, and the resulting p-values are summarized in Table \ref{tab:stat_tests_pvalues}. These results show that the differences between linear probing and \ac{LoRA} are statistically significant for all three datasets ($p \ll 0.001$), with \ac{LoRA} consistently outperforming linear probing. Likewise, baseline models significantly outperform linear probing for all datasets ($p \ll 0.0001$). 

We also note that the results of EfficientNet on the AMi-Br test dataset is better than reported in previous work~\citep{bertram2025histologic}. This can be attributed to the fact that the latter used EfficientNet V2-S, while we use EfficientNet V2-M in our experiments.


We also notice a drop in performance for several settings, including both end-to-end fine-tuned models and foundation models (both linear probing and \acp{LoRA}-fine-tuned) in the two external test datasets, as reflected by the balanced accuracy and AUROC scores. We attribute this to the domain shift between the datasets. For the AtNorM-Br dataset, the observed performance may partially also be influenced by the reliance on a single expert involved in the classification of \acp{AMF}, potentially introducing systematic label biases and limiting generalizability, an effect that evens out when multiple raters are involved. Moreover, an image distribution shift cannot be ruled out, as the images were retrieved from different sources (labs, scanners, etc.), making the dataset more heterogeneous. For the AtNorM-MD dataset, where five annotators were involved, a systematic label bias is rather unlikely, however, the inclusion of multiple domains beyond breast cancer introduces a more pronounced image domain shift, encompassing a wider range of tissue types that pose serious challenges for model adaptation.

A significant limitation of our approach lies in training exclusively on the AMi-Br dataset with a limited number of cases of only human breast cancer, restricting the models' ability to learn the full spectrum of \ac{AMF} representations across different tissue types, tumors, institutions, staining protocols, and scanners. Our results emphasize that automated \ac{AMF} identification remains a highly challenging problem in computational pathology, requiring further methodological advances. Our new, multi-domain datasets, provided by this work, help further this cause. Future work will address the integration of these classifier into a two-stage mitotic figure detection approach, and investigate the role of false positive / negative detections.

\begin{acronym}
\acro{EMA}[EMA]{exponential moving average}
\acro{HE}[H\&E]{Hematoxylin \& Eosin}
\acro{ROI}[RoI]{region of interest}
\acro{WSI}[WSI]{whole slide image}
\acro{MF}[MF]{mitotic figure}
\acro{TCGA}[TCGA]{The Cancer Genome Atlas}
\acro{CMC}[CMC]{canine mammary carcinoma}
\acro{AtNorM}[AtNorM]{Atypical and Normal Mitosis}
\acro{AMF}[AMF]{atypical mitotic figure}
\acro{CNN}[CNN]{convolutional neural network}
\acro{ViT}[ViT]{vision transformer}
\acro{LoRA}[LoRA]{Low Rank Adaptation}
\acro{ROC}[ROC]{receiver operating characteristic}
\acro{AUROC}[AUROC]{Area under the Receiver Operating Characteristic Curve}
\end{acronym}


\section*{Author contributions}

SB wrote the main manuscript and carried out the experiments. VW, TAD, TC, RK, and CAB served as experts in the dataset annotation. SB and JA carried out the statistical analysis. CAB, KB, and MA co-wrote the manuscript and guided the project. RHJF organized and compiled the AtNorM-Br dataset with the support of CAB as pathology expert. CK provided scientific expertise. NP created Figure 1. 
All authors reviewed the manuscript.

\acks{CAB, VW, and CK acknowledge the support from the Austrian Research Fund (FWF, project number: I 6555). SB, TC, RK, and MA acknowledge support by the Deutsche Forschungsgemeinschaft (DFG, German Research Foundation,
project number: 520330054). KB acknowledges support by the German Research Foundation (DFG) project 460333672 CRC1540 EBM. JA acknowledges support by the Bavarian State Ministry of the Sciences and the Arts (project FOKUS-TML). NP and MA acknowledge support by the Deutsche Forschungsgemeinschaft (DFG, German Research Foundation, project number: 545049923)
}

%
\ethics{The work follows appropriate ethical standards in conducting research and writing the manuscript, following all applicable laws and regulations regarding treatment of animals or human subjects.}

\coi{We declare no conflicts of interest.}


\begin{sloppypar}
\data{All data used in this study is public and can be found at  \url{https://github.com/DeepMicroscopy/AMi-Br_Benchmark}.}
\end{sloppypar}

\bibliography{sample}


\end{document}